Prediction of adverse events in Afghanistan: regression analysis of time series data grouped not by geographic dependencies.

Krzysztof Fiok 1, Waldemar Karwowski 1, Maciej Wilamowski 2
1 University of Central Florida, Department of Industrial Engineering and Management Systems, Orlando, Florida, USA
2 University of Warsaw, Faculty of Economic Sciences, Warsaw, Poland

Abstract

The aim of this study was to approach a difficult regression task on highly unbalanced data regarding active theater of war in Afghanistan. Our focus was set on predicting the negative events number without distinguishing precise nature of the events given historical data on investment and negative events per each of predefined 400 Afghanistan's districts. In contrast with previous research on the matter, we propose an approach to analysis of time series data that benefits from non-conventional aggregation of these territorial entities. By carrying out initial exploratory data analysis we demonstrate that dividing data according to our proposal allows to identify strong trend and seasonal components in the selected target variable. Utilizing this approach we also tried to estimate which data regarding investments is most important for prediction performance. Based on our exploratory analysis and previous research we prepared 5 sets of independent variables that were fed to 3 machine learning regression models. The results expressed by mean absolute and mean square errors indicate that leveraging historical data regarding target variable allows for reasonable performance, however unfortunately other proposed independent variables doesn't seem to improve prediction quality.

Introduction

In any conflict the ability to predict negative events before they take place gives edge over the adversary. Leveraging modern machine learning techniques enables to analyze historical conflict-related data and deliver high-quality predictions which is why these methods are very often applied. In this study we aimed to apply machine learning methods to a dataset containing data from Afghanistan which was introduced an analyzed in previous studies (1,2,3,4). The authors of these studies utilized various approaches and models in order to solve various regression and binary classification tasks formulated on the data. (1,2,3) addressed regression tasks with 4 dependent variables, namely number of dead, wounded, hijacked personnel and overall number of negative events. In (1,2) various fuzzy inference systems were evaluated, in (3) shallow artificial neural networks and multiple linear regression models were employed and in (4) for solving of binary classification task 5 model types were used, namely, neural networks, k-nearest neighbors, support vector machines, random forests and C4.5 decision trees. Since the discussed dataset reflects administrative division of territorial entities in Afghanistan, all above mentioned studies followed the same manner of aggregating data into districts, provinces and regions. Such approach resulted in difficult to solve problem of sparse target variable vectors e.g. (4) calculated that vector of negative events target variable for whole Afghanistan is in 87.23% filled with "0" values. In our work we proposed a method of overcoming this difficulty by different aggregation of districts and training selected machine learning models separately for the new resulting groups.

Methods

Dataset and Exploratory Data Analysis (EDA)

This study analyzes data described in (1,2,3,4) and addresses a regression task for one target variable, namely amount of negative events (NE) without distinguishing between types of NE i.e. dead, wounded or hijacked personnel. Inspired by observation by (4) that this target variable is a sparse vector, we focused on tackling this problem. Our analysis approaches the entire data set comprising of 33 600 data points as a time series with 400 realizations individual for each district, with time step of 1 month and duration of 7 years, resulting in 84 data points for each district realization. At first we tried to aggregate the realizations in a natural geographically-constrained manner as in (1,2,3,4), namely in groups of provinces and regions also provided in the dataset. To initially explore data grouped in this manner we computed the all-time per district average negative events value (ANE) and concluded that realizations in these geographic groups varied strongly. This seamed challenging for any prediction model and was in agreement with conclusions by (1,2,3,4), so we searched for another approach to grouping 400 districts. After creating ANE histogram presented in figure 1, we found that all realizations can be divided by ANE into 3 groups: A with ANE = 0, B with ANE in range (0, 2> and C with ANE greater than 2. It appeared that in group A there are 140 districts, 247 districts fall into group B and 13 districts belong to group C. It can be concluded, that 140 out of total 400 districts are "silent" i.e. have always NE = 0 value, regardless of external factors. Predicting always 0 as the number of NE for districts from group A is a perfect solution and doesn't require any prediction model. In contrast, groups B and C are characterized by percentage of data points with NE = 0 of 83.49 % for group B and 20.97 % for group C. This shows that the proposed groups differ strongly and in fact can be treated as different data generating processes that should be modeled separately. Therefore, our aim was to model two separate time series for group B and C separately.

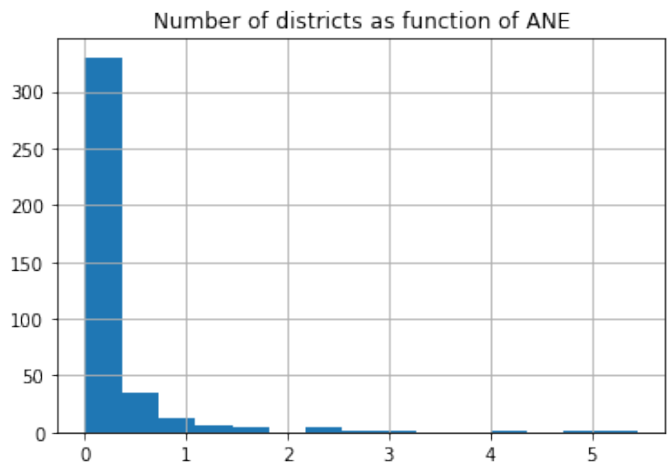

Figure 1. Number of districts as function of all-time per district average negative events value (ANE)

Time series decomposition

As part of EDA we decided to create two separate time series, namely for group B and separately group C by computing a Month-by-Month Average Negative Event value for all districts (MMANE) in the group (MMANE B and MMANE C). Further, we decomposed these time series in a manner popular e.g. in economic sciences into trend, seasonal and residual components (https://otexts.com/fpp2/classical-decomposition.html). We compared additive, multiplicative and STL (Cleveland et al., 1990) decomposition methods by the mean absolute residual component synonymous with mean absolute error (MAE) computed for each method. The results are presented in table 1.

Table 1. MAE for tested decomposition methods for MMANE B and MMANE C time series

| Group | Average MMANE | MAE with various decomposition methods | | |
|---|---|---|---|---|
| | | Additive | Multiplicative | STL |
| B | 0.270 | 0.082 | 1.531 | 0.099 |
| C | 3.443 | 0.776 | 3.440 | 0.816 |

Figure 2. Curves derived from additive decomposition method for MMANE B and MMANE C time

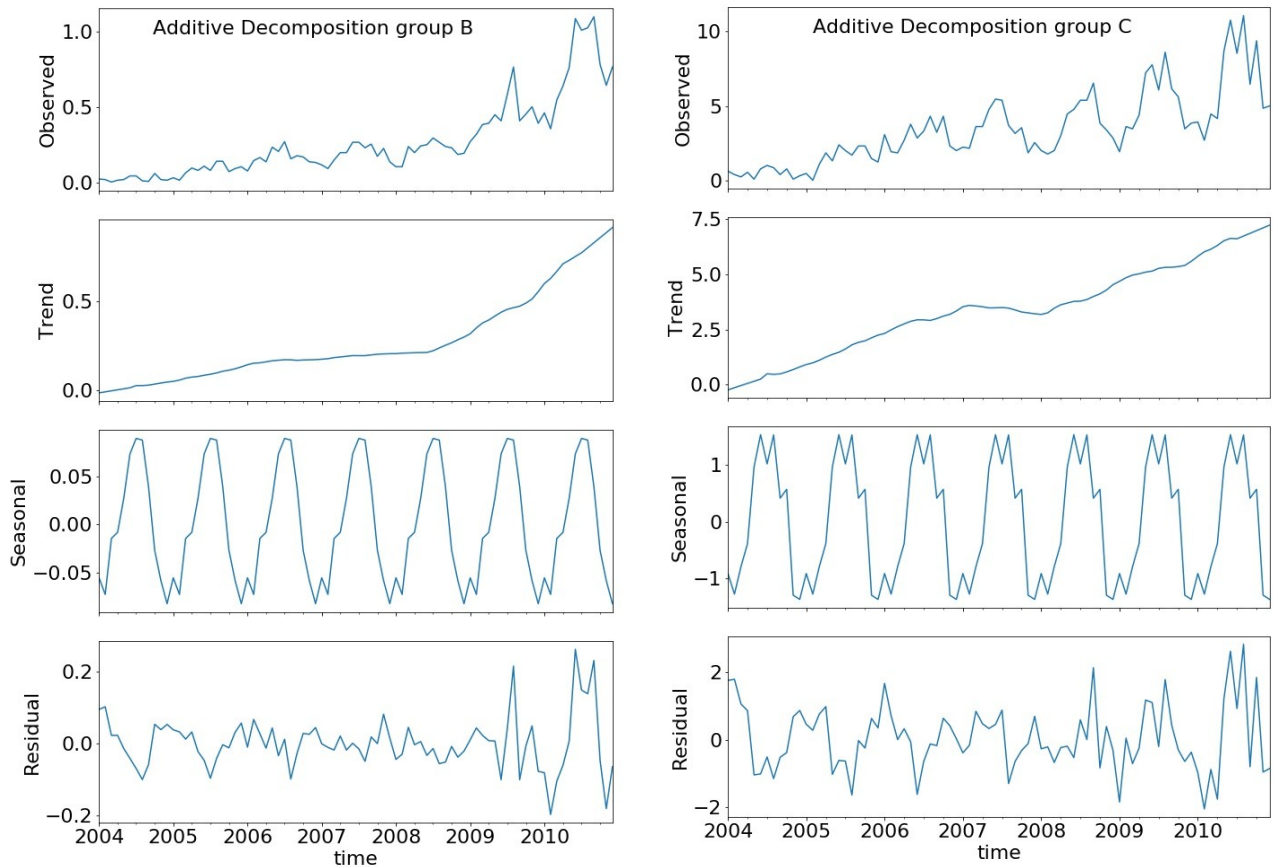

series.

Based on figure 2, it can be concluded that in both groups all curves have somewhat similar shapes indicating obvious increase of MMANE over years and strong seasonality over months.

It can be concluded that leveraging basic time-related historic information regarding MMANE allows to carry out prediction of MMANE with MAE of 0.082 when addressing districts from group B and 0.776 when dealing with group C districts. Given average MMANE for group B and C was 0.270 and

3.443 respectively, this translates to average prediction error of 30.27 % for group B and 22.54 % for group C.

To preliminary explore importance of the investment data present in the dataset, for all investment types, budget amounts and number of investment projects we followed the Distributed Lag (DL) approach proposed by (7) and computed information regarding past month-by-month values in range <-1, -12> months. We further employed a basic linear regression model (LRM) to try explaining the residual component left after additive decomposition with use of investment information.

Unfortunately, in both district groups B and C each realization was a time series composed of only 84 data points. In order to for the model no to over-fit to abundance of provided investment-related input variables we fed only one single lagged variable at a time e.g. A3 investment type lagged 4 months or B5 investment budget lagged 8 months. This allowed us to compare the MAE decrease in each case. In most cases the decrease of MAE was close to none and the lagged variables that provided best MAE decrease for group B and C are presented in table 2. It can be observed that the most valuable investment information for districts from group B was the number of emergency assistance projects carried out 6 months backwards (coded as A6-6) which allowed for over 5 % improvement in MAE. Analogically, for districts from group C improvement of over 6 % MAE was achieved with use of information regarding amount of investments in gender carried out 4 months backwards (coded as B9-4).

Table 2. MAE for additive decomposition and linear regression model, information that allowed best performance and percentage gain achieved by leveraging investment information.

| Group | Additive decomposition | Linear Regression Model | Information | Percentage Difference |
|---|---|---|---|---|
| B | 0.082 | 0.078 | A6-6 | 5.319 |
| C | 0.776 | 0.730 | B9-4 | 6.020 |

Preparation of data and features fed to machine learning (ML) models

Firstly, we tested auto regression (AR) models fed only with historical data regarding target variable. Since EDA demonstrated that there is a strong trend and seasonal component in historic information regarding MMANE, for each data point we prepared the following time related information (TRI) derived from original data:
- NE value for the analyzed district for -m, where m is months in range <-1,-12>;
- 3 month trend of MMANE value for all districts understood as: MMANE(-3m) - MMANE(-1m);
- MMANE value for all districts for last year;
- MMANE value for all districts for last half year;
- MMANE value for all districts for last 3 months;
- MMANE value for all districts for last 1 month;
- month of data instance; and
- year of data instance.

Secondly, we made use of the investment-related information that was prepared as previously described with use of DL approach. In addition, before feeding other data to machine learning models we made use of basic feature engineering methods.

Similarly as in previously described LRM approach, in order for the models not to over-fit to the data, we tried to keep low the number of independent variable columns and considered only selected information regarding investments (SID), namely A6-6 and B9-4 i.e. the two features that were found most informative during EDA. Also, because previous research (4) found the information regarding number of Community Development projects carried out a year before the data instance (coded as A2(t-1)) important for predicting number of NE, we also considered this information. In the very foundation of our concept for analyzing the discussed data lies importance of district identification number (ID), therefore we added also this feature to analysis. In order to compare influence of selected information on ML model performance, we decided to define 5 variants of feature sets fed to ML models:

V1) include only TRI;
V2) include TRI + district identification number (ID)
V3) include TRI + selected investment data (SID)
V4) include TRI + ID + SID
V5) include TRI + ID + SID + A2(t-1)

Therefore, models trained on V1 and V2 feature sets can be treated as AR models, whereas models trained on V3, V4 and V5 feature sets followed autoregressive distributed lag approach proposed in (6).

Time series data split

It is possible to analyze the dataset as a whole, namely, taken into consideration that it is comprised of time series for each district, divide the data in a way that years 2004-2009 are used for training and 2010 for testing. However, apart from this approach, we also proposed splitting the data in another manner. Since the data comprised of 7 years of observations, we propose to split it to 6 ML tasks that progressively address longer and longer time period. Additionally, because the last year witnessed dramatically increased number of NE when compared to earlier years, we added a separate 8th data split that uses only 2 last years of observations. The whole pattern applied for creating 7 train-test data splits is presented in table 3. Taken into consideration that we already divided the data into group B and group C, this constitutes 14 ML tasks. Also, including the fact that we defined 5 different feature sets, results in overall 70 separate ML tasks.

Table 3. Data split for 7 machine learning tasks.

| ML task | Year | | | | | | |
|---|---|---|---|---|---|---|---|
| | 2004 | 2005 | 2006 | 2007 | 2008 | 2009 | 2010 |
| T1 | Train | Test | | | | | |
| T2 | Train | Train | Test | | | | |
| T3 | Train | Train | Train | Test | | | |
| T4 | Train | Train | Train | Train | Test | | |
| T5 | Train | Train | Train | Train | Train | Test | |
| T6 | Train | Train | Train | Train | Train | Train | Test |
| T7 | | | | | | Train | Test |

Selected machine learning models, metrics and visualization techniques

For the regression task we selected to compare performance of 3 machine learning models in each of 70 defined tasks: linear regression (LR) as baseline, random forest (RF) and gradient boosting (GB). As a result, we ended up comparing performance of 210 ML model variants. For each defined ML task we also computed metrics for a naive baseline model predicting always 0 as the NE value.

We adopted two metrics for the addressed regression tasks, namely mean absolute error (MAE) and mean square error (MSE). To visually assess model performance we created plots visualizing directly the predictions made by chosen ML models in chosen ML tasks. Also, we benefit from the fact that the random forest ML models during training develop ranks that reflect influence of independent variables on the target variable. We used these ranks to create visualizations of feature importance for all ML tasks in the V5 variant which includes all analyzed features.

Software

Data preparation and ML models were implemented in Python with use of numpy, pandas, statsmodels and sci-kit learn packages. All computations were carried out on the same computing machine.

Results

Districts group B

Table 4. MSE and MAE in districts from group B, calculated for all ML models in all variants and ML tasks. "0" columns represent a model predicting always "0". Best models in a given ML task according to a given metric are highlighted with bold font.

|  | MSE | | | | MAE | | | |
| --- | --- | --- | --- | --- | --- | --- | --- | --- |
| ML Task | LR | GB | RF | 0 | LR | GB | RF | 0 |
| T1V1 | 0.2576 | 0.1608 | **0.1493** | 0.1532 | 0.3148 | 0.1151 | 0.1212 | **0.0864** |
| T1V2 | 0.2576 | 0.1670 | **0.1463** | 0.1532 | 0.3148 | 0.1176 | 0.1209 | **0.0864** |
| T1V3 | 0.2512 | 0.1626 | **0.1444** | 0.1532 | 0.3080 | 0.1144 | 0.1153 | **0.0864** |
| T1V4 | 0.2512 | 0.1622 | **0.1463** | 0.1532 | 0.3080 | 0.1142 | 0.1153 | **0.0864** |
| T1V5 | 0.2793 | 0.1603 | **0.1474** | 0.1532 | 0.3364 | 0.1129 | 0.1165 | **0.0864** |
| T2V1 | **0.2119** | 0.3031 | 0.2382 | 0.2861 | 0.2008 | 0.2581 | 0.2339 | **0.1680** |
| T2V2 | **0.2119** | 0.2802 | 0.2386 | 0.2861 | 0.2009 | 0.2444 | 0.2359 | **0.1680** |
| T2V3 | **0.2118** | 0.3211 | 0.2395 | 0.2861 | 0.2011 | 0.2872 | 0.2509 | **0.1680** |
| T2V4 | **0.2118** | 0.2706 | 0.2411 | 0.2861 | 0.2011 | 0.2515 | 0.2487 | **0.1680** |
| T2V5 | **0.2117** | 0.2709 | 0.2467 | 0.2861 | 0.2198 | 0.2578 | 0.2914 | **0.1680** |
| T3V1 | **0.2451** | 0.2507 | 0.2529 | 0.3408 | 0.3041 | 0.2709 | 0.2914 | **0.1930** |
| T3V2 | **0.2451** | 0.2501 | 0.2521 | 0.3408 | 0.3042 | 0.2693 | 0.2901 | **0.1930** |
| T3V3 | **0.2450** | 0.2511 | 0.2544 | 0.3408 | 0.3036 | 0.2715 | 0.2948 | **0.1930** |
| T3V4 | **0.2451** | 0.2490 | 0.2508 | 0.3408 | 0.3037 | 0.2654 | 0.2918 | **0.1930** |
| T3V5 | **0.2446** | 0.2761 | 0.2615 | 0.3408 | 0.3003 | 0.3130 | 0.3251 | **0.1930** |
| T4V1 | **0.2466** | 0.2636 | 0.2682 | 0.3816 | 0.2946 | 0.2836 | 0.2965 | **0.2129** |
| T4V2 | **0.2466** | 0.2653 | 0.2722 | 0.3816 | 0.2946 | 0.2876 | 0.3057 | **0.2129** |
| T4V3 | **0.2466** | 0.2621 | 0.2706 | 0.3816 | 0.2943 | 0.2890 | 0.2996 | **0.2129** |
| T4V4 | **0.2466** | 0.2637 | 0.2742 | 0.3816 | 0.2943 | 0.2853 | 0.3094 | **0.2129** |
| T4V5 | **0.2508** | 0.2637 | 0.2778 | 0.3816 | 0.2729 | 0.2827 | 0.3370 | **0.2129** |
| T5V1 | **0.6447** | 0.6774 | 0.6491 | 1.0557 | 0.4671 | 0.4686 | 0.5004 | **0.4443** |
| T5V2 | **0.6447** | 0.6600 | 0.6576 | 1.0557 | 0.4671 | 0.4679 | 0.5083 | **0.4443** |
| T5V3 | **0.6445** | 0.6754 | 0.6485 | 1.0557 | 0.4677 | 0.4687 | 0.5024 | **0.4443** |
| T5V4 | **0.6445** | 0.6676 | 0.6637 | 1.0557 | 0.4677 | 0.4680 | 0.5095 | **0.4443** |
| T5V5 | **0.6416** | 0.6558 | 0.6543 | 1.0557 | 0.4752 | 0.4713 | 0.5140 | **0.4443** |
| T6V1 | **1.2481** | 1.3389 | 1.2905 | 2.5617 | **0.6733** | 0.7080 | 0.7081 | 0.7642 |
| T6V2 | **1.2483** | 1.3688 | 1.3119 | 2.5617 | **0.6733** | 0.7169 | 0.7157 | 0.7642 |
| T6V3 | **1.2478** | 1.3126 | 1.2988 | 2.5617 | **0.6738** | 0.7047 | 0.7081 | 0.7642 |
| T6V4 | **1.2479** | 1.3679 | 1.3165 | 2.5617 | **0.6738** | 0.7157 | 0.7176 | 0.7642 |
| T6V5 | **1.2496** | 1.3403 | 1.3177 | 2.5617 | **0.6710** | 0.7056 | 0.7137 | 0.7642 |
| T7V1 | 16.5333 | 1.3876 | **1.3029** | 2.5617 | 3.8593 | 0.7128 | **0.6947** | 0.7642 |
| T7V2 | 16.5335 | 1.3765 | **1.3206** | 2.5617 | 3.8593 | **0.6895** | 0.7041 | 0.7642 |
| T7V3 | 16.5179 | 1.3810 | **1.2957** | 2.5617 | 3.8573 | 0.7055 | **0.6953** | 0.7642 |
| T7V4 | 16.5179 | 1.3755 | **1.3334** | 2.5617 | 3.8573 | **0.6993** | 0.7046 | 0.7642 |
| T7V5 | 16.4298 | 1.3750 | **1.3038** | 2.5617 | 3.8456 | 0.7072 | **0.7028** | 0.7642 |

Districts group C

Table 5. MSE and MAE in districts from group C, calculated for all ML models in all variants and ML tasks. "0" columns represent a model predicting always "0". Best models in a given ML task according to a given metric are highlighted with bold font.

|  | MSE | | | | MAE | | | |
|---|---|---|---|---|---|---|---|---|
| ML Task | LR | GB | RF | 0 | LR | GB | RF | 0 |
| T1V1 | 4.9629 | 4.3942 | **3.7332** | 5.6987 | 1.5336 | 1.3718 | **1.2832** | 1.5064 |
| T1V2 | 4.8230 | 4.1628 | **3.7691** | 5.6987 | 1.5070 | 1.3542 | **1.2929** | 1.5064 |
| T1V3 | 4.7400 | 4.1208 | **3.8153** | 5.6987 | 1.5089 | 1.3141 | **1.3059** | 1.5064 |
| T1V4 | 4.5946 | 4.0779 | **3.8292** | 5.6987 | 1.4864 | 1.3412 | **1.3119** | 1.5064 |
| T1V5 | 7.1238 | 4.2611 | **3.8117** | 5.6987 | 1.9265 | 1.3590 | **1.3073** | 1.5064 |
| T2V1 | 15.0127 | 7.1098 | **5.5800** | 14.9167 | 3.1722 | 1.9985 | **1.7329** | 2.9679 |
| T2V2 | 14.9904 | 6.2262 | **5.6207** | 14.9167 | 3.1690 | 1.8687 | **1.7294** | 2.9679 |
| T2V3 | 15.4418 | 6.8422 | **5.6481** | 14.9167 | 3.2380 | 1.9675 | **1.7506** | 2.9679 |
| T2V4 | 15.4129 | 6.2497 | **5.7372** | 14.9167 | 3.2337 | 1.8722 | **1.7676** | 2.9679 |
| T2V5 | 12.9999 | 6.5810 | **5.4852** | 14.9167 | 2.8919 | 1.9427 | **1.7153** | 2.9679 |
| T3V1 | **6.3327** | 7.6324 | 6.5986 | 20.6667 | **1.9940** | 2.1213 | 2.0167 | 3.5000 |
| T3V2 | **6.3192** | 7.4348 | 6.4088 | 20.6667 | **1.9958** | 2.1203 | 2.0088 | 3.5000 |
| T3V3 | **6.3307** | 7.6014 | 6.4211 | 20.6667 | **1.9956** | 2.1919 | 2.0076 | 3.5000 |
| T3V4 | **6.3158** | 7.4159 | 6.3804 | 20.6667 | **1.9965** | 2.1268 | 2.0118 | 3.5000 |
| T3V5 | **6.3301** | 7.7172 | 6.6047 | 20.6667 | **1.9913** | 2.2077 | 2.0373 | 3.5000 |
| T4V1 | 8.8548 | 7.0326 | **6.5218** | 24.4103 | 2.5682 | 2.0294 | **1.9845** | 3.7821 |
| T4V2 | 8.8038 | 6.6956 | **6.3028** | 24.4103 | 2.5603 | 1.9751 | **1.9222** | 3.7821 |
| T4V3 | 8.9570 | 7.0816 | **6.4488** | 24.4103 | 2.5838 | 2.0208 | **1.9641** | 3.7821 |
| T4V4 | 8.9103 | 6.4253 | **6.3484** | 24.4103 | 2.5765 | 1.9282 | **1.9287** | 3.7821 |
| T4V5 | 8.8064 | 6.8129 | **6.2429** | 24.4103 | 2.5592 | 1.9791 | **1.9241** | 3.7821 |
| T5V1 | **7.4709** | 8.5805 | 7.4951 | 39.0256 | 2.1647 | 2.2763 | **2.1370** | 5.1795 |
| T5V2 | **7.5482** | 9.2017 | 7.5835 | 39.0256 | 2.1801 | 2.3049 | **2.1342** | 5.1795 |
| T5V3 | 7.4558 | 8.4634 | **7.4038** | 39.0256 | 2.1632 | 2.2609 | **2.1104** | 5.1795 |
| T5V4 | **7.5352** | 9.1743 | 7.6810 | 39.0256 | 2.1782 | 2.2919 | **2.1715** | 5.1795 |
| T5V5 | **7.5240** | 9.2207 | 7.9961 | 39.0256 | **2.1816** | 2.2893 | 2.2051 | 5.1795 |
| T6V1 | **19.6499** | 23.3164 | 23.8695 | 76.8590 | **2.9930** | 3.3036 | 3.1765 | 6.6667 |
| T6V2 | **19.6847** | 23.5193 | 23.9741 | 76.8590 | **2.9937** | 3.2342 | 3.1384 | 6.6667 |
| T6V3 | **19.6534** | 23.8503 | 24.1001 | 76.8590 | **2.9933** | 3.3027 | 3.1693 | 6.6667 |
| T6V4 | **19.6878** | 23.0254 | 23.8466 | 76.8590 | **2.9941** | 3.2276 | 3.1699 | 6.6667 |
| T6V5 | **19.6378** | 23.2192 | 24.1943 | 76.8590 | **2.9972** | 3.2007 | 3.1957 | 6.6667 |
| T7V1 | 942.2008 | 25.6683 | **24.3162** | 76.8590 | 30.1682 | 3.3580 | **3.1583** | 6.6667 |
| T7V2 | 932.8895 | 28.3220 | **25.2844** | 76.8590 | 30.0135 | 3.5611 | **3.2220** | 6.6667 |
| T7V3 | 956.8620 | 25.7500 | **24.7452** | 76.8590 | 30.4067 | 3.3764 | **3.1692** | 6.6667 |
| T7V4 | 936.8120 | 28.0776 | **25.0899** | 76.8590 | 30.0779 | 3.5697 | **3.1958** | 6.6667 |
| T7V5 | 953.8071 | 27.6032 | **24.6550** | 76.8590 | 30.3515 | 3.4947 | **3.1952** | 6.6667 |

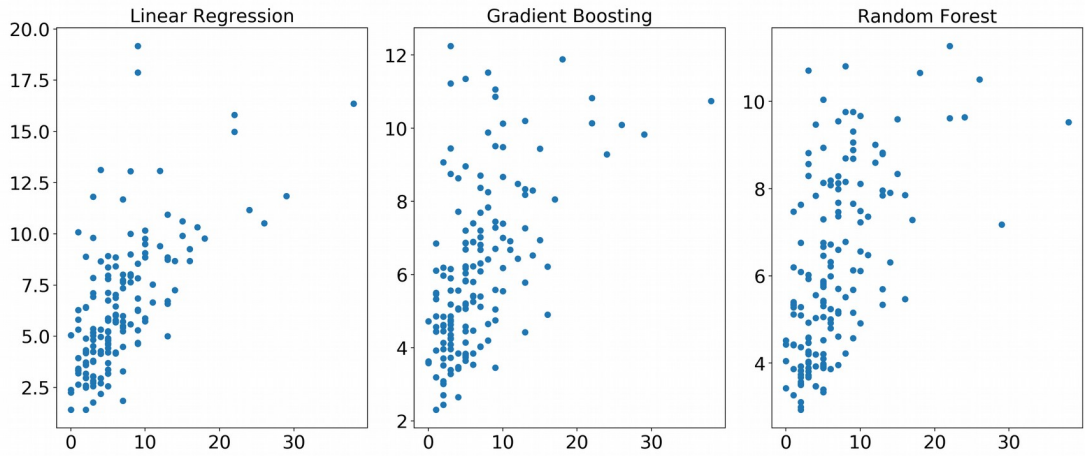

Figure 3. From left to right: predictions of LR, GB and RF models in ML task T6V5 for districts from group C

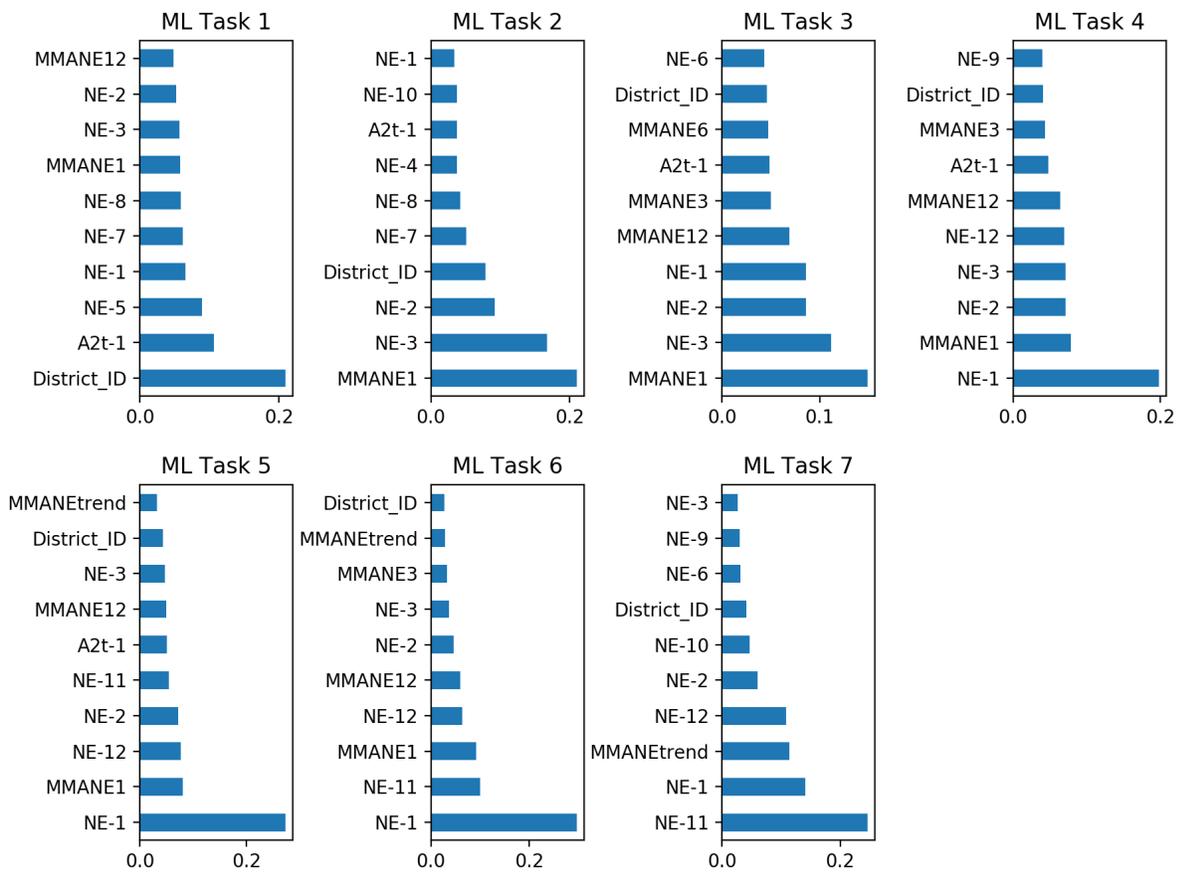

Figure 4. Feature importance in districts from group C. From left to right: ML task V5T1-T7

Discussion

The mean square error (MSE) and mean absolute error (MAE) are smaller for group B districts when compared to group C districts, which is an obvious conclusion provided the all-time per district average negative events value (ANE) value for these groups is 0.270 and 3.443 respectively.

Closer analysis of results for group B with MAE regarded as the quality measure shows that the strategy "to predict always 0" as the number of negative events is best compared to evaluated models in ML tasks T1-T5. Most likely this can be attributed to low number of negative events in these tasks. Only in tasks 6 and 7, which address years richer in negative events, other models come to play. When considering MSE, the "0" strategy does not bring satisfactory results at all and the 3 analyzed ML models outperform it by a strong margin. However discussing which of these models performs best is difficult since differences in errors are vague. One conclusion that can be drawn is, that linear regression performs poorly in T1 and T7, that is in cases where least training data is provided.

More pronounced differences between models performance are visible in results of group C districts. Here, due to numerous negative events, the "0 strategy" is never preferred. The smallest errors, both MSE and MAE, are achieved either by Random Forest or Linear Regression with Gradient Boosting following closely. Figure 3 presents examples of predictions in district group C carried out by 3 analyzed models with use of all proposed features and in a machine learning task where all years from available data where used. The shapes created by the plotted predictions are similar and only with precisely described errors in table 5 one can conclude which model is superior to others. This confirms, that the models perform similarly.

An insight into informativeness of features fed to ML models is provided by figure 4. There we are able to compare 10 most important features according to Random Forest regression model in each ML task and in the variant of analysis that uses all proposed features. It can be observed, that in tasks T1 to T5 there is one investment feature that seems important for predictions, namely A2(t-1) as proposed in (4). District ID is mentioned in all tasks and in T1, where little historical information is present, it seems to be the most important feature. From the time related information features (TRI) the prevailing ones are MMANE-1, NE-1 and NE-11. This indicates that the number of negative events from previous month in the given district as well as averaged over districts, play an important role. However before jumping into conclusions it must be noted, that for each ML task the MSE and MAE are extremely similar regardless of feature set fed to the ML models. For instance in the analyzed ML task T6 in all feature set variants V1-5 the average and standard deviation of MAE achieved by Random Forest model is 3.1699 ± 0.0185, therefore standard deviation is only 0.58 % of the average value. This shows there is very small influence of changes in the feature set on model performance. At the same time this puts a question mark on the usefulness of carried out exploratory data analysis regarding the choice of most informative investment-related features and on overall possibility of extracting information that could improve prediction quality from available investment data.

Conclusions

Dividing the dataset into groups of districts according to the all-time per district average negative event value (ANE) allows to simplify the machine learning task by instantly excluding from analysis 140 out of 400 districts with no negative events. The group of districts with non 0 but still small number of negative events constitutes a very difficult regression task, where in many cases all evaluated models are outperformed by "always predict 0" strategy. Only with the rise of number of negative events it is possible to demonstrate superiority of machine learning models over naive approach. In this case, our analysis found that the discussed time series data has strong trend and visible seasonal component. As a result reasonable quality of predictions can be achieved simply with use of historical data regarding negative events. Adding features regarding investments or providing district identification number explicitly to the trained models doesn't seem to bring measurable improvement in performance.